\documentclass[letterpaper, 10 pt, conference]{ieeeconf}  % Comment this line out if you need a4paper

\IEEEoverridecommandlockouts                              % This command is only needed if 
\usepackage[utf8]{inputenc}
\usepackage[T1]{fontenc}
\usepackage{graphicx}
\usepackage{tabularx}
\usepackage{float}
\usepackage{bm}
\usepackage[space]{cite}
\usepackage{algpseudocode}
\usepackage{algorithm}
\usepackage{hyperref}
\usepackage{flushend}
\usepackage{amsmath} % assumes amsmath package installed
\usepackage{color}

\title{\LARGE \bf FF-SRL: High Performance GPU-Based Surgical Simulation For Robot Learning}

\author{Diego Dall'Alba$^{1,2,*}$, Micha{\l} Naskr{\k e}t$^{1,*}$, Sabina Kamińska$^{1}$ and Przemys{\l}aw Korzeniowski$^{1,\dagger}$% <-this % stops a space
\thanks{The publication was created within the project of the Minister of Science and Higher Education "Support for the activity of Centers of Excellence established in Poland under Horizon 2020" on the basis of the contract number MEiN/2023/DIR/3796. This project has received funding from the EU’s Horizon 2020 research and innovation programme under grant agreement No 857533. This publication is supported by Sano project carried out within the International Research Agendas programme of the Foundation for Polish Science, co-financed by the EU under the European Regional Development Fund.  We acknowledge Polish high-performance computing infrastructure PLGrid (HPC Center: ACK Cyfronet AGH) for providing computer facilities and support within the grant no. PLG/2020/013635}
\thanks{* These authors contributed equally to the paper. Ordered alphabetically.}
\thanks{$^{1}$ Sano Centre for Computational Medicine, Kraków, Poland {https://sano.science/}}
\thanks{$^{2}$ Department of Engineering for Innovation Medicine, University of Verona, Italy}
\thanks{$\dagger$ Corresponding author: Przemys{\l}aw Korzeniowski (email: p.korzeniowski@sanoscience.org)}
}

\begin{document}

\maketitle
\thispagestyle{empty}
\pagestyle{empty}

%%%%%%%%%%%%%%%%%%%%%%%%%%%%%%%%%%%%%%%%%%%%%%%%%%%%%%%%%%%%%%%%%%%%%%%%%%%%%%%%
\begin{abstract}
%Autonomous robotic surgery is a promising field that can benefit from the automation of surgical tasks using Reinforcement Learning (RL) techniques. However, RL training requires a high number of trials, which are often unsafe or impractical to perform on real systems. Therefore, simulation environments are essential for RL training, but they need to be realistic, scalable, and efficient. 
%In this work, we propose FF-SRL, an entirely GPU-based simulation environment for robotic surgery that supports the simulation of deformable tissue via an extended position based dynamics model. We integrate FF-SRL into a fully GPU RL training pipeline and compare it with other available simulators on the task of reaching a deformable tissue. 
%Our results show that FF-SRL can significantly improve the performance of RL training, reducing the training time from tens of minutes to less than two minutes. We make our code publicly available to foster the development of autonomous surgical systems using RL techniques.

Robotic surgery is a rapidly developing field that can greatly benefit from the automation of surgical tasks.
However, training techniques such as Reinforcement Learning (RL) require a high number of task repetitions, which are generally unsafe and impractical to perform on real surgical systems.
This stresses the need for simulated surgical environments, which are not only realistic, but also computationally efficient and scalable.
We introduce FF-SRL (Fast and Flexible Surgical Reinforcement Learning), a high-performance learning environment for robotic surgery. In FF-SRL both physics simulation and RL policy training reside entirely on a single GPU. This avoids typical bottlenecks associated with data transfer between the CPU and GPU, leading to accelerated learning rates.
Our results show that FF-SRL reduces the training time of a complex tissue manipulation task by an order of magnitude, down to a couple of minutes, compared to a common CPU/GPU simulator.
Such speed-up may facilitate the experimentation with RL techniques and contribute to the development of new generation of surgical systems. To this end, we make our code publicly available to the community.

%%%%%%%%%%%%%ORIGINAL TEXT FROM IROS2023, WE NEED TO REWRITE IT
% Reinforcement Learning (RL) algorithms applied to robots present an opportunity to be implemented in a wide field of applications. One that is of high importance is healthcare where robots might be able to support surgeons in a variety of visuomotor tasks. To this end, creating robust virtual environments representing the real clinical cases will serve as resource-optimal playgrounds.

% In this work we present a Position Based Dynamics (PBD) based suite that allows for creating massively parallel virtual environments in surgical context. Using the cutting-edge software and hardware solutions we were able to confine computations to Graphics Processing Units (GPU) which significantly reduces simulation processing times. %Do we have time to actually test it on cpu?

% The solution we are presenting can be later on applied to RL algorithms in order to create robotic policies that will be transferable to real world cases (Sim-To-Real).

\end{abstract}

%%%%%%%%%%%%%%%%%%%%%%%%%%%%%%%%%%%%%%%%%%%%%%%%%%%%%%%%%%%%%%%%%%%%%%%%%%%%%%%%
\section{INTRODUCTION}
Robot-Assisted Surgical Systems (RASS) are becoming more popular and widely used, with a constant increase in their adoption in the last two decades \cite{d2021_10years}. Researchers are also exploring new possibilities and challenges of RASS, using platforms like da Vinci Research Kit (dVRK) \cite{kazanzides2014dvrk}.

One of the most interesting research topics is how to automate some aspects of RASS interventions, making these systems more autonomous and efficient \cite{fiorini2022concepts}. A common approach is to use Reinforcement Learning (RL) techniques \cite{ richter2019dvrl,  xu2021surrol, varier2022ambf-rl, tagliabue2020unityflexml, scheikl2023lapgym}, which have proven to be effective in other robotic domains, such as legged locomotion and dexterous manipulation \cite{makoviychuk2021isaac}. RL can support RASS in performing various surgical tasks that require adaptability and precision, such as manipulation of deformable tissues \cite{tagliabue2020unityflexml,Sheikl_SimToReal_2022} or suture needle \cite{chiu2021NMbimanual,schwaner2021LFD_NM} and removing debris or liquids \cite{richter2019dvrl,xu2021surrol, varier2022ambf-rl}.

RL-based systems are usually trained in simulated environments, as many unsuccessful attempts are required which often exhibit unsafe behavior \cite{makoviychuk2021isaac}. This is especially true for RASS, since performing real experiments faces significant economic and ethical constraints. However, to transfer the trained models in real settings, it is necessary to employ realistic, scalable, and robust simulations, supporting soft-body physics \cite{tagliabue2020unityflexml, Sheikl_SimToReal_2022}.
% \begin{figure}[t]
%     \centering
%     \includegraphics[width=0.8\columnwidth]{images/singleEnvB.png}
%     \caption{An example frame of tissue retraction procedure simulation created within our framework, including a model of RASS instrument (orange shaft and red gripper), deformable fat tissue (yellow) and rigid supporting base (dark gray) including a target area of interest (purple).}
%     \label{fig:senv}
% \end{figure}
 \begin{figure}[t]
    \centering
    \includegraphics[width=\columnwidth]{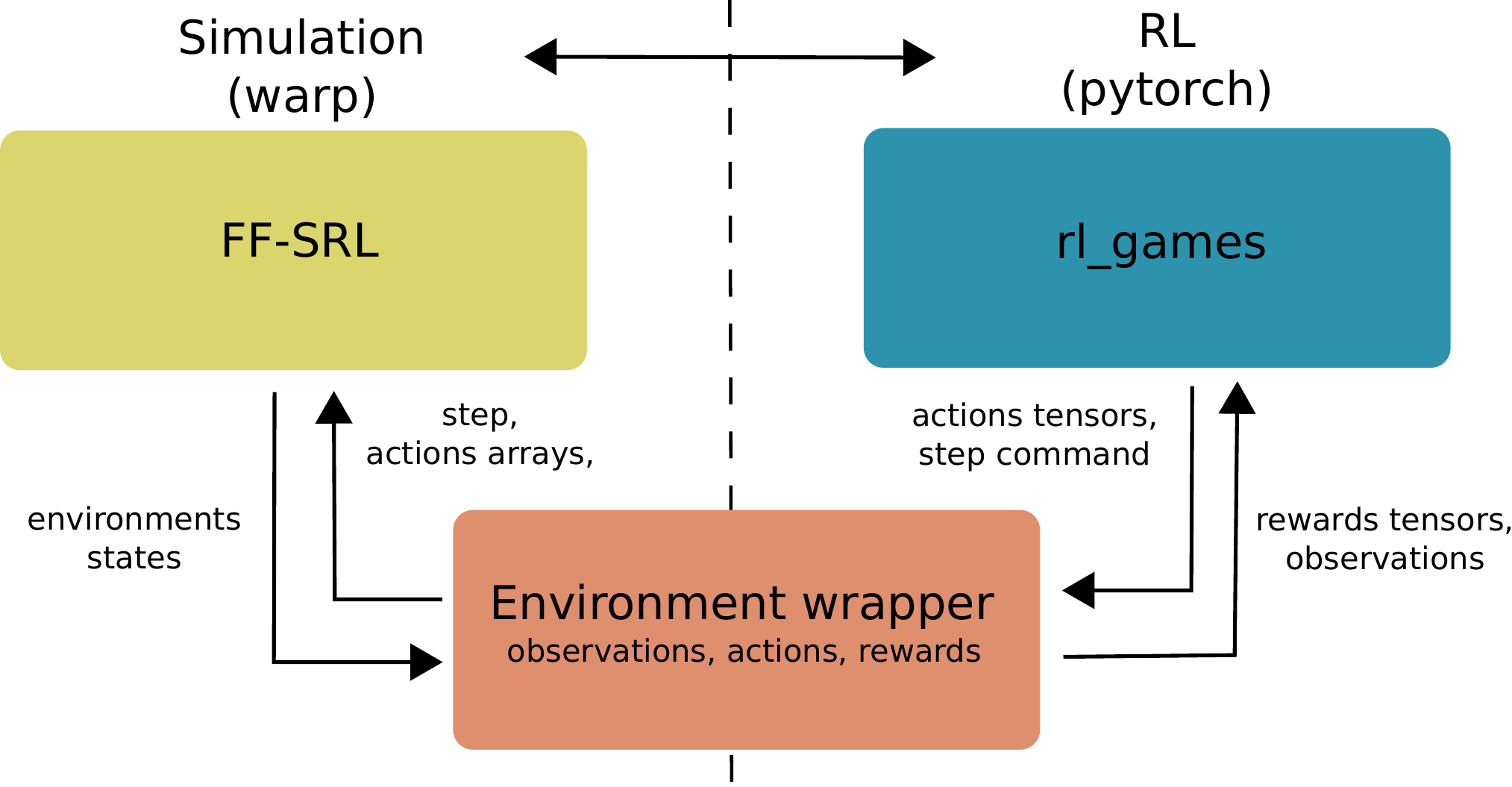}
    \caption{Schematic representation of the integration between the proposed FF-SRL simulator and the rl\_games RL framework to achieve a fully GPU training pipeline and support the development of autonomous RASS. }
    \label{fig:pipeline}
\end{figure}
Several simulators for RASS have been proposed, focusing on specific aspects, such as the kinematic model of the robot \cite{richter2019dvrl}, more realistic physical \cite{scheikl2023lapgym} or visual \cite{xu2021surrol} simulation or easier interfacing with real RASS \cite{varier2022ambf-rl}.
All these simulators mainly exploit CPU calculations, relegating the use of the GPU to accelerate the training and inference part of the RL model only. 

This is not optimal as it requires copying the data back and forth between CPU (simulation) and GPU (learning) memories . As such, it limits the complexity of the tasks that can be solved and, as a result, slows down research activities and community contributions. The solution is to compute the entire simulation and RL training process on the GPU. However, developing a massively-parallel GPU simulation is non-trivial. Its efficient implementation  requires a careful handling of thousands of simultaneous threads and efficient global memory access patterns.

In this work we therefore propose:
\begin{itemize}
    %\item a simulation environment for surgical robotics, called FF-SRL, that runs entirely on GPU and support soft tissue modeling via an eXtended Position Base Dynamics (XPBD) approach;
    \item a high-performance, fully GPU-based simulator for surgical robotics, called FF-SRL, that supports soft tissue modeling via extended position based dynamics approach;
    \item the integration of FF-SRL into an entirely GPU-based RL training pipeline;
    \item the experimental validation of the proposed FF-SRL in a fundamental surgical task (i.e. reaching a deformable tissue), comparing with simulator proposed in \cite{scheikl2023lapgym}.
\end{itemize}

%The work considers a tissue retraction task which has already been demonstrated to be able to be transferred effectively between simulator and reality~\cite{tagliabue2020unityflexml, Sheikl_SimToReal_2022}. Tests with the real robot will be carried out in the future to demonstrate the capabilities of the proposed simulator to also interface with real robotic systems, exploiting ROS middleware.

To the best of our knowledge, this is the first work that proposes a simulation environment for training autonomous RASS with RL techniques entirely on GPU, and the results confirm that our approach is able to significantly improve the performance compared to available RASS simulators, e.g. completing RL training in less than 2 minutes instead of over 1 hour. This enables training of complex learning tasks even on a commonly accessible consumer-grade hardware such as gaming laptops.

We make the simulator code publicly available\footnote{ \url{https://github.com/SanoScience/FF-SRL}} to encourage the community to adopt our approach and accelerate the progress of this promising research area.

\section{RELATED WORKS} \label{sec:sota}
Research activities on RASS have been exploring the automation of various surgical tasks: these tasks range from simple ones, such as peg transfer \cite{hu2023peg,meli2020peg,hwang2020peg_superhuman}, 
%or autonomous endoscope movements \cite{pandya2014reviewECM, gao2022ECMsavanet}, 
to more complex ones, such as %cutting patterns \cite{nguyen2019CUTnew,thananjeyan2017CUTmultilateral},
manipulating suture needles \cite{chiu2021NMbimanual,schwaner2021LFD_NM} or deformable tissues \cite{meli2021TRAutonomous,tagliabue2022TRAdeliberation, d2022TROlearning, tagliabue2021TROdata}. Among these tasks, we focus on tissue retraction, a preliminary phase of most surgical interventions. Tissue retraction involves lifting deformable tissue to expose an area of interest, such as an organ or lesion to be removed (see Fig.~\ref{fig:move}). 
%This task is simple to set up but challenging enough to test the proposed automation approaches, and that is suited to be learned in a simulated environment and subsequently verified in a real environment \cite{tagliabue2020unityflexml,Sheikl_SimToReal_2022,pore2021LFD}.
This task is widely used by the RASS research community since it is simple to set up but, at the same time, with an adequate level of complexity for testing the developed automation approaches. Furthermore, this task can be learned in simulation and transferred to a real environment \cite{tagliabue2020unityflexml,Sheikl_SimToReal_2022,pore2021LFD}.

As task complexity has increased, there has been a growing interest in applying more complex automation approaches, such as RL techniques.
RL allows the inclusion of multi-agent models \cite{scheikl2021TROcooperative}, safety constraints \cite{pore2021TROsafe}, or demonstrations by expert users \cite{schwaner2021LFD_NM, pore2021LFD, huang2023LFDguided}, also through interactive supervision \cite{long2023interactive}. To make RL training more effective, it is a common practice to carry it out in realistic simulation environments.
Several simulation environments have been proposed for RASS. Some of them are based on rigid objects only, such as dVRL \cite{richter2019dvrl} and SurRoL \cite{xu2021surrol}. dVRL provides an accurate kinematic model of the dVRK platform and simulates simple training or surgical tasks. SurRoL provides around ten surgical tasks supported by a more realistic visual rendering. Another simulator is AMBF-RL \cite{varier2022ambf-rl}, which extends Articulated Multi-Body Framework (AMBF) \cite{munawar2019ambf} simulator specific for RL training and ensures simple interfacing with the real setup.
However, none of these simulators exploit the simulation of deformable objects, which are essential for many surgical tasks. UnityFlexML \cite{tagliabue2020unityflexml} was the first simulator to address this issue by taking advantage of the Position-Based Dynamics (PBD) approach provided by NVIDIA Flex and integrating it within the Unity framework. LapGym \cite{scheikl2023lapgym} recently proposed a more advanced simulation environment, which relies on the accurate biomechanical finite element simulation provided by SOFA \cite{Faure2012, schegg2023sofagym}. Compared to other simulators, LapGym implements a rich catalog of surgical tasks, most based on interaction with deformable objects.

Thus, we propose a novel simulator that offers an entirely GPU-integrated RL simulation and training approach for RASS. Unlike described simulators that combine CPU calculations with a limited GPU-accelerated part, FF-SRL leverages the full power of GPU computing to optimize the overall process. To tackle the complexity of the simulation of highly non-linear behavior of tissues FF-SRL uses eXtended Postion-Based Dynamics (XPBD \cite{MacklinXPBD2016}). Several recent papers \cite{MacklinSmallSteps2019, MacklinContact2020, MullerRigidBody2020, macklin2021constraint, mueller2022shapeMatching}, turned XPBD  into a serious competitor of more advanced simulation methods in terms of accuracy, stability, speed, and simplicity.

\section{METHODS}
In this section, we describe the proposed simulation environment, detailing the implementation choices regarding soft tissue simulation, RASS tools and collisions. Next, we describe how we included support for RL environments. We point out that some implementation choices have sacrificed pure computational performance in favor of a software architecture that simplifies maintenance and extensibility.
% \subsection{Software framework}

% The simulation is developed within the NVIDIA Warp framework which allows for all simulation computations to be done on GPU. Warp is created in python which makes it user friendly and easy to use. In Warp arrays of data stored on GPU are conveniently created in a similar way to PyTorch and NumPy libraries. The arrays can be copied between CPU and GPU on demand and Warp implements helper functions to move them from and to PyTorch tensors. One of the features implemented recently is capability to run kernels on multiple CUDA devices which can be used to further improve parallelization possibilities on powerful devices such as NVIDIA A-100.

\subsection{Simulation framework}
FF-SRL implements an XPBD simulation technique~\cite{MacklinXPBD2016}, based on NVIDIA Warp \cite{warp2022}, which is a Python framework for writing high-performance GPU-optimized simulation and graphics code. While Warp provides an XPBD implementation, we have implemented a version specifically optimized to handle the complexity of RASS simulation, including non-linear tissue behavior.

%In order to deal with the complexity of simulating RASS scenario, including non-linear tissue behavior, we utilized XPBD as our simulation technique~\cite{MacklinXPBD2016}. Our implementation is based on NVIDIA Warp \cite{warp2022}, which is a Python framework for writing high-performance GPU-optimized simulation and graphics code.
%PBD has advantages over global matrix-based solvers due to the local nature of the non-linear Jacobi process, allowing for stable handling of equality and inequality constraints. 
%While Warp provides an implementation of the XPBD approach, it is intended for general applications and is not suitable for simulating complex deformable objects.
%We use an extended implementation of the PBD method (XPBD), to improve the effectiveness of constraint solver iterations in simulating complex deformable objects. 

In particular, the proposed XPBD implementation improves the effectiveness of constraint solver iterations in simulating complex deformable objects. We split each time step into $n$ smaller sub-steps and apply a single constraint iteration of PBD in each sub-step. This method, first suggested in~\cite{MacklinSmallSteps2019}, is more effective than performing a single large time-step with $n$ constraint solver iterations. This approach has been further validated in~\cite{MullerRigidBody2020}, where a comprehensive rigid body solver was implemented to handle various joint types and the interaction with soft objects in a unified way. This is particularly useful in RASS, allowing easy interaction between rigid surgical tools and soft tissues. The authors of~\cite{MacklinXPBD2016,MullerRigidBody2020} compared PBD with sub-stepping to more advanced solvers, both implicit and explicit, and found that it produces visually similar results while maintaining the simplicity of the original PBD method and reducing the sensitivity to matrix ill-conditioning. 
%Recent studies have demonstrated that PBD is an accurate, stable, fast, and simple competitor to more advanced methods ~\cite{MacklinSmallSteps2019, MullerRigidBody2020}.

\subsubsection{Implementation of soft-body physics}

Each object is represented by a set of $V$ vertices with masses $m_{i}$, where $i=\{1, \cdots, V\}$, a position $\bm{x_i}$ and velocity $\bm{v_i}$. Vertices are embedded within a volumetric tetrahedral mesh and the objects' surface is divided into $F$ triangle faces. Two adjacent vertices are combined into an edge. Two sets of constraints are used to provide object deformability with minimal computation required -- conserving distance between points $C^{\text{d}}$ and volume of tetrahedrons $C^{\text{v}}$. Each constraint imposes corrections $\Delta\bm{x}_i$ on corresponding vertices' positions. %For clarity we skip the numbering of vertices in the equations that will follow.

The distance constraint $C^\text{d}$ is realized by the following corrections for vertices $a, b$ residing on a given edge:
\begin{align}
    \begin{split}
        \Delta \bm{x_a^\text{d}} = - k_\text{s} \frac{w_{a}}{w_{a} + w_{b}} \left( |\bm{x_{a,b}}| - d_{a,b}\right) \frac{\bm{x_{a,b}}}{|\bm{x_{a,b}}|},\\
        \Delta \bm{x_b^\text{d}} = + k_\text{s} \frac{w_{b}}{w_{a} + w_{b}} \left( |\bm{x_{a,b}}| - d_{a,b}\right) \frac{\bm{x_{a,b}}}{|\bm{x_{a,b}}|},
    \end{split}
\end{align}
where $\bm{x_{a,b}}=\bm{x_a} - \bm{x_b}$, $w_{a}$, $w_{b}$ are inverse masses of the vertices $a$ and $b$, $d_{a,b}$ is the rest distance between vertices $a, b$ calculated before the simulation loop, and $k_\text{s} \in [0,1]$ is the constraint stiffness.

The volume constraint $C^\text{v}=\frac{1}{6}(\bm{x_{b,a}}\times\bm{x_{c,a}})\cdot\bm{x_{d,a}} - V_0$ (where $V_0$ is the rest volume of the tetrahedron) is calculated by the following corrections for the vertices  $a, b, c, d$ residing on a given tetrahedron:
\begin{align}
    \begin{split}
        \Delta \bm{x_a^\text{v}} =& - \frac{1}{6} s k_\text{v} \cdot\\
        &(\bm{x_{b,a}}\times\bm{x_{c,a}} + \bm{x_{c,a}}\times\bm{x_{d,a}} + \bm{x_{d,a}}\times\bm{x_{b,a}}),\\
        \Delta \bm{x_b^\text{v}} =& - \frac{1}{6} s k_\text{v} (\bm{x_{b,a}}\times\bm{x_{c,a}}),\\
        \Delta \bm{x_c^\text{v}} =& - \frac{1}{6} s k_\text{v} (\bm{x_{c,a}}\times\bm{x_{d,a}}),\\
        \Delta \bm{x_d^\text{v}} =& - \frac{1}{6} s k_\text{v} (\bm{x_{b,a}}\times\bm{x_{d,b}}),
    \end{split}
\end{align}
where $k_\text{v}$ is the stiffness and $s$ is scaling factor:
\begin{align}
    \begin{split}
        s = \frac{C^\text{v}}{\sum_{i\in\{a,b,c,d\}}|\nabla_{\bm{x}_i} C^\text{v}|^2}.
    \end{split}
\end{align}
We parallelized the algorithm using Jacobi-style constraint solver and atomic add function. In this case all the constraints can be calculated in parallel and atomically added as a vertex' total correction. Subsequently, the correction is averaged by the number of constraints applied to the vertex to obtain the final position correction $\overline{\Delta\bm{x}} = \Delta\bm{x}/{n}$. The stiffness parameters $k_\text{s}$ and $k_\text{v}$ are adopted from previous studies~\cite{tagliabue2020unityflexml, meli2021TRAutonomous, tagliabue2021TROdata}. 

% \begin{align}
%     \begin{split}
%         \overline{\Delta\bm{x}} = \frac{\Delta\bm{x}}{n}.
%     \end{split}
% \end{align}

% Przemek, please check
Tissue connections between different objects are implemented as distance constraints between adjacent vertices of one mesh and geometrical centers of triangles of other mesh. The constraint is applied with position correction $\Delta\bm{x}^\text{c}$.

An overview of the algorithm loop is presented in Algorithm~\ref{alg:simulation} with $\bm{g}$ being the gravity constant.
\begin{algorithm} [htb]
    \caption{Simulation algorithm.}
    \label{alg:simulation}
    \begin{algorithmic}[1]
        \State set initial positions $\bm{x_0}$ and velocities $\bm{v_0}$
        \State $h \gets \Delta t/ simSubSteps$
        \State $k \gets 0$
        \While{$i < simSteps$}
            \State evaluate instrument state
            \While{$j < simSubSteps$}
            \State predict velocity: $\Tilde{\bm{v}} \gets \bm{v}_{k} + h \bm{g}$
            \State predict position: $\Tilde{\bm{x}} \gets \bm{x}_{k} + h \Tilde{\bm{v}}$
            \State initialize corrections: $\Delta \Tilde{\bm{x}} \gets 0$
                \ForAll{vertices}
                    \State $\Delta \Tilde{\bm{x}} \gets \Delta \Tilde{\bm{x}} + \Delta \Tilde{\bm{x}}^\text{d} + \Delta \Tilde{\bm{x}}^\text{g} + \Delta \Tilde{\bm{x}}^\text{c}$
                \EndFor
                \ForAll{tetrahedra}    
                    \State $\Delta \Tilde{\bm{x}} \gets \Delta \Tilde{\bm{x}} + \Delta \Tilde{\bm{x}}^\text{v}$
                \EndFor
                \State $\Tilde{\bm{x}} \gets \Tilde{\bm{x}} + \Delta \Tilde{\bm{x}}/n$
                \State $\bm{v}_{k+1} \gets (\Tilde{\bm{x}} - \bm{x}_k)/h$
                \State $\bm{x}_{k+1} \gets \Tilde{\bm{x}}$
                \State $k \gets k+1$
                \State $j \gets j+1$
            \EndWhile
            \State evaluate collisions
            \State $i \gets i+1$
        \EndWhile
    \end{algorithmic}
\end{algorithm}

\subsubsection{Surgical tool model} \label{sec:tool}
%A simple model of laparoscopic grasper is employed as it is not a crucial feature to check the simulation performance. 
The surgical tool model employs two layers: visual and computational. The former is only a graphical representation of instrument mesh and the latter is used for collision detection. 
%As presented in Fig.~\ref{fig:grasper} 
The computational model consists of three capsules: one acting as a rod corresponding to instrument's shaft, and two acting as clamps. The clamps are nested at the end of the rod and can pivot around their fixed point with angle $0^{\circ}<\alpha<30^{\circ}$ from their symmetry axis. 

%The instrument has 3 degrees of freedom that correspond to Cartesian coordinates movement. 
We considered a simplified model of the movement of the instruments, but still representative of those used in minimally invasive surgical procedures, including RASS ones \cite{scheikl2023lapgym, richter2019dvrl}. We did not consider the distal wrist joints in this initial implementation, since most RL work in RASS does not control the orientation of the tool, keeping it constant \cite{tagliabue2020unityflexml, pore2021LFD, pore2021TROsafe, Sheikl_SimToReal_2022}. The main constraint that must be satisfied is imposed by the access port (i.e. the trocar) necessary to pass through the patient's abdominal wall. Therefore, the tool can rotate freely with respect to the access point $\bm{p_{RCM}}$ and translate along the shaft axis \cite{scheikl2023lapgym}. However, the control of the instrument is commonly carried out in Cartesian space, so we have implemented a simple mapping of the instrument's movements. 

\begin{figure}[tp]
    \centering
    \includegraphics[width=\columnwidth]{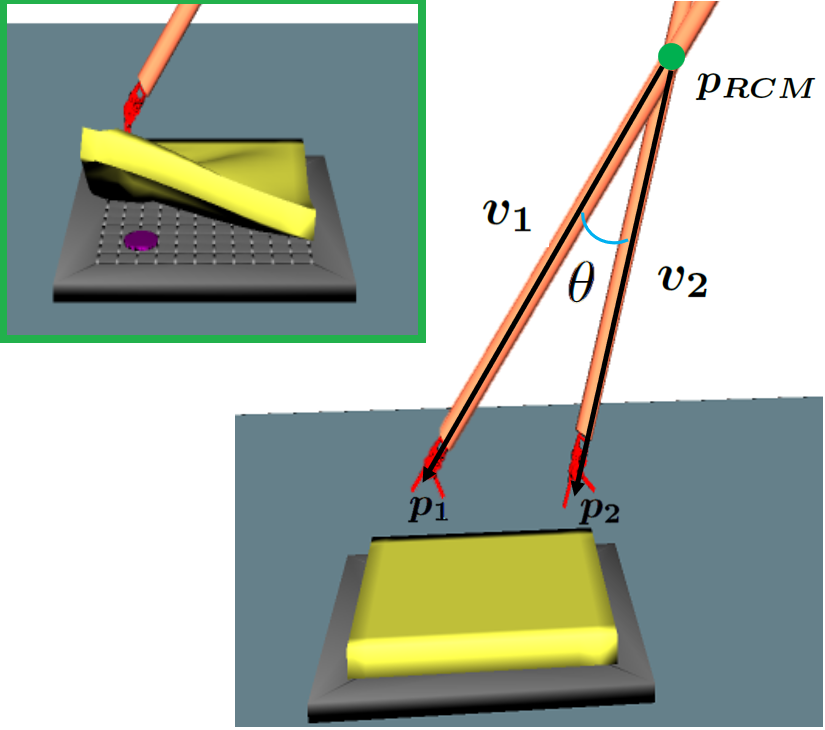}
    \caption{Representation of the instrument movement model, required to ensure compliance with the remote center of motion; see the main text for further details. In the subfigure outlined in green, we show an example frame of tissue retraction procedure simulation created within our framework, including a model of the RASS instrument (orange shaft and red gripper), deformable fat tissue (yellow), and rigid supporting base (dark gray), including a target area of interest (purple).}
    \label{fig:move}
\end{figure}

As represented in Fig.~\ref{fig:move}, given the instrument in two different poses to which the distal points (center of the grasper) $\bm{p_1}$ and $\bm{p_2}$ correspond. Given the position of the constraint point $\bm{p_{RCM}}$, we can calculate the vectors $\bm{v_1} = \bm{p_1} - \bm{p_{RCM}}$ and  $\bm{v_2} = \bm{p_2} - \bm{p_{RCM}} $. Given these vectors it is possible to calculate the rotation angle $\theta$ and the respective axis $\bm{r_A}$ with the formulas:
\begin{align}
    \theta = \cos^{-1}\left(\frac{\bm{v_1} \cdot \bm{v_2}}{||\bm{v_1}||\cdot ||\bm{v_2}||}\right), \quad
    \bm{r_A} = \frac{\bm{v_1} \times \bm{v_2}}{||\bm{v_1}||\cdot||\bm{v_2}||} 
\end{align}
%$\theta = \cos^{-1}(\frac{\bm{v_1} \cdot \bm{v_2}}{||\bm{v_1}||\cdot ||\bm{v_2}||}) $ and $\bm{r_A} = \frac{\bm{v_1} \times \bm{v_2}}{||\bm{v_1}||\cdot||\bm{v_2}||} $. 
These rotations satisfy the  $\bm{p_{RCM}}$ constraint, while the translation component can be obtained from $|| \bm{v_1} - \bm{v_2} ||$.

%The displacement of the laparoscopic grasper is provided to the simulation in form of $3$-element vectors of floating point numbers each frame, as for the simple task tested it is not necessary to include grasping information.
% \begin{figure}[h]
%     \centering
%     \includegraphics[width=0.45\columnwidth]{images/laparoscope.pdf}
%     \caption{A simple laparoscopic grasper model is used with rod capsule marked in blue and clamp capsules marked with green. The grasping and clamp pivot points are marked with cyan and magenta points, respectively.}
%     \label{fig:grasper}
% \end{figure}

The surgical tool model has a predetermined dragging point positioned at the end of closed clamps (see points $\bm{p_1}$ and $\bm{p_2}$ in Fig.~\ref{fig:move}). When the angle between clamps is lower than $3^{\circ}$ the algorithm checks for the closest vertex in radius $r_l$. If a vertex is found in the radius $r_l$ then a dragging constraint is applied to the mesh and the dragging point. When the angle between clamps exceeds $3^{\circ}$ all the dragging constraints are deactivated.
Dragging simulation is carried out as creating an additional distance constraint with $k_\text{s} = 1$ between the dragging point and a mesh vertex. The constraint is applied with position correction $\Delta\bm{x}^\text{g}$. All vertices have an additional variable that stores the boolean value of the dragging constraint state.

\subsubsection{Collisions}
 In order to provide a robust solution for collision detection between mesh triangles and the surgical tool model we employ Signed Distance Field (SDF) approach. SDFs are a popular choice for the collision detection shapes due to their performance \cite{tagliabue2021TROdata}. We follow the per-element local optimisation scheme to find intersection points between SDFs and continuous surfaces proposed in~\cite{MacklinContact2020}. Each capsule of the grasper model is treated as a SDF and its penetration is checked against the mesh' faces.

 For each intersecting face an additional distance constraint is introduced, that is applied after deformation constraints are calculated. The constraint pushes the face out of the grasper SDFs in the direction of penetration vector and proportionally to the penetration depth with stiffness of $k_\text{c} = 1$.

%  \begin{figure}[tb]
%     \centering
%     \includegraphics[width=\columnwidth]{images/pipelineV2.pdf}
%     \caption{Schematic representation of the integration between the FF-SRL simulator and the rl\_games RL framework to achieve a fully GPU training pipeline and support the development of autonomous RASS systems. }
%     \label{fig:pipeline}
% \end{figure}
% \begin{figure}[t]
%     \centering
%     \includegraphics[width=0.8\columnwidth]{images/singleEnvB.png}
%     \caption{An example frame of tissue retraction procedure simulation created within our framework, including a model of RASS instrument (orange shaft and red gripper), deformable fat tissue (yellow) and rigid supporting base (dark gray) including a target area of interest (purple).}
%     \label{fig:senv}
% \end{figure}

\subsection{Reinforcement learning integration} \label{sec:rl_games}
The developed simulator was integrated with the rl\_games framework, which provides highly optimized end-to-end GPU implementation of several RL algorithms \cite{rl-games2021}. These implementations vectorize observations and actions on the GPU allowing to fully exploit the parallelization provided by the developed simulator. As represented in Fig.~\ref{fig:pipeline}, an efficient wrapping module has been implemented to support the simulator which takes care of transferring data between the simulator and the RL algorithm, taking advantage of the primitives provided by Warp for interfacing with PyTorch framework used by rl\_games. The latter allow for CUDA interoperability between Warp arrays and PyTorch tensors without copying the underlying data.

The interface between the developed simulator and rl\_games package is also used to calculate, transform and transfer the necessary data for the RL process. Namely, episode rewards are calculated and scaled, observations are extracted from the simulation, and actions are applied when calling the simulation's step function once per simulation step. The interface also checks for termination conditions, i.e. the limit on the number of steps and task fulfillment conditions.

%In particular, the state of the environment is transferred to the RL agent in the form of observation vector and reward, while the actions generated by the agent are applied in the simulator. 
%Each simulation environment is stored in an independent vector, which includes, in addition to the vectors required to represent the objects in the scene (e.g. vertices, faces, tetrahedra), also the tensors containing the information for the RL agent.
For each variable required in the calculations (e.g. vertices, faces, tetrahedrons and tensors containing the information for the RL agent) one GPU array is created and data from each environment is concatenated and stored in the array. 
% This is presented schematically in Fig.~\ref{fig:memory}.

% \begin{figure}[htb]
%     \centering
%     \includegraphics[width=0.85\columnwidth]{images/memeory.pdf}
%     \caption{The data for all parallel environments is stored in the same GPU array.}
%     \label{fig:memory}
% \end{figure}

\section{EXPERIMENTS}

The experiments have the dual objective of evaluating the performance of the FF-SRL simulator alone and the integration into the RL training process. We are considering tissue retraction since it is a widely used task in validating autonomous RASS, as described in Section \ref{sec:sota}. The environment, presented in Fig.~\ref{fig:move}, replicates the characteristics of the one proposed in \cite{scheikl2023lapgym, Sheikl_SimToReal_2022, tagliabue2021TROdata, tagliabue2022TRAdeliberation}. In addition to the RASS tool, there is a camera with the point of view shown in the figure, a rigid base in which an area of interest is presented and to which the soft tissue is fixed along the side furthest from the camera. The movement of the tool is implemented in Cartesian space, as described in Section \ref{sec:tool}. %\todo{MISSING DETAILS?}
The proposed method is compared with LapGym \cite{scheikl2023lapgym} since it provides a reference implementation of the tissue retraction task with a realistic CPU simulation of the deformable tissue based on a finite element model.

\begin{table}[tb]
\centering
\caption{Hyperparameters of PPO used for the experiments. The steps before PPO update are summed over all environments and $lin(x)$ is a linear schedule starting at $x$ and ending at $0$}
\label{tab:hyperparameters}
\setlength\tabcolsep{2pt}
\begin{tabular*}{\columnwidth}{@{\extracolsep{\fill}} c|c||c|c }
%\begin{tabular}{c|c || c|c}
    Hyperparameter & Value & Hyperparameter & Value\\
\hline
    Total simulation steps & $5\cdot10^{5}$ & Minibatch size & $256$\\
    Steps before update & $1024$ & Update epochs & $4$\\ 
    Discount factor $\gamma$ & $0.995$ & $\lambda_\text{GAE}$ & $0.95$\\
    Clip range value func. & $0.2$ & Clip range & $lin(0.1)$\\
    Value function coeff. & $0.5$ & Entropy coeff. & $0.0$\\
    Max. gradient norm & $0.5$ & Learning rate & $lin(2.5\cdot10^{-4})$
\end{tabular*}
\end{table}
%We initially evaluate only the simulation performance and scaling capabilities of the proposed simulator in terms of environments running in parallel, considering the fps generated by the simulator without any RL training process running. For each environment, $n_{iter} = 10000$ simulation steps are performed. At the beginning of each episode, a random vector of three actions per environment is generated and used to execute the movement of the surgical instrument.
\begin{table*}[!htb]
%\centering
\caption{Frames per second as the number of environments running in parallel (left) or the complexity of the bio-mechanical model (right) increases for the simulation environments considered. The reported values are in the form mean $\pm$ standard deviation calculated over 5 runs initialized with random seeds.}
\label{tab:fps}
\begin{tabular}{c|cc|cc }
 Number of &   \multicolumn{2}{c|}{Simulation Only}           & \multicolumn{2}{c}{RL Setup}    \\
Enviroments & FF-SRL        & LapGym       & FF-SRL        & LapGym  \\
\hline
1      & $2964.6\pm40.6$     & $77.7\pm0.7$  & $434.3\pm8$    & $68\pm1.1$  \\
4      & $9093.2\pm132.8$   & $205.8\pm7.2$ & $1527.3\pm34$  & $173.8\pm8.4$ \\
8      & $14021.3\pm193.4$ & $226.1\pm11.5$  & $2636.7\pm40.7$  & $192.3\pm9.9$  \\
16     & $19032.9\pm98.1$   &  --          & $3955.4\pm202.1$  &  --           \\
32     & $21662.2\pm145.3$   &  --          & $4992\pm335.7$ &  --            \\
64     & $23728.7\pm314$   &   --         &  --               &   --          \\
80     & $23932.7\pm317.4$  &   --          &  --               &   --         
\end{tabular}
\quad\quad\quad
\begin{tabular}{c|cc}
% \setlength\tabcolsep{2pt}
% \begin{tabular*}{\columnwidth}{@{\extracolsep{\fill}} c|cc}
Number of &   \multicolumn{2}{c}{Simulation Only}  \\
 Tetrahedral & FF-SRL        & LapGym      \\
\hline
1170      & $2964.6\pm40.6$     & $77.7\pm0.7$   \\
1431      & $2798.3\pm118$   & $73.6\pm2.1$  \\
2880      & $2751.9\pm119.3$ & $48.2\pm1.3$   \\
9729     & $2302.2\pm44.5$   & $37.7\pm0.4$        \\
52359   & $905.5\pm29$   &  $9.6\pm0.1$ \\
\multicolumn{2}{c}{} & \\
\multicolumn{2}{c}{} & 
\end{tabular}
\end{table*}

We initially evaluate only the simulation performance and scaling capabilities of the proposed simulator in terms of environments running in parallel and as the complexity of the biomechanical model of the simulated tissue increases (obtained by increasing the number of vertices of the tissue model). For each environment, we consider the fps generated from $n_{iter} = 500000$ simulation steps performed without any RL training process running. The measurements do not consider initialization times to ensure a fair comparison with the data obtained during RL training \cite{makoviychuk2021isaac}.

We then evaluate the FF-SRL simulator applied to RL training, considering the first phase of the tissue retraction task, which consists of reaching a point on the surface of the soft tissue. This choice is consistent with the approach in \cite{varier2022ambf-rl} and replicates the task of reaching anatomical areas of interest which is fundamental for most surgical activities, as adopted by \cite{xu2021surrol,richter2019dvrl}. In this task, the instrument starts from a fixed point above the tissue and reaches a fixed target point on the tissue surface.
The training is based on the Proximal Policy Optimization (PPO) algorithm \cite{Schulman2017PPO} since it is considered a suitable baseline, given its wide successful applications in heterogeneous fields \cite{makoviychuk2021isaac}, including RASS \cite{tagliabue2020unityflexml, pore2021TROsafe, scheikl2023lapgym, Sheikl_SimToReal_2022}.
However, the simulation conforms to OpenAI \textit{gym}~\cite{Brockman2016OpenAI} standard and can therefore be applied to other RL algorithms and libraries in a straightforward way.
%The training is based on the Proximal Policy Optimization (PPO) algorithm \cite{Schulman2017PPO}, given that it is an algorithm widely used successfully in various fields\cite{andrychowicz2020dexterous,mirhoseini2021chip}, including RASS \cite{tagliabue2020unityflexml, pore2021TROsafe, scheikl2023lapgym, Sheikl_SimToReal_2022}.
%using the same parameters proposed in \cite{scheikl2023lapgym}. 

Although FF-SRL supports the use of generated endoscopic images, we consider state observations that contain the position of the instrument and the target point to be reached, combined with the reward $R = w_{l}l + w_{d}d + w_{s}s$ ;
where $l$ is the distance between the end-effector and the target, $d$ is the change of $l$ with respect to previous simulation step and $s$ is a success flag normally at 0 and set to 1 when the distance $l < 3$ mm. The weights $w_{l},w_{d},w_{s}$ are set to -1, -10 and 100, respectively.

% \begin{table}[htb]
% \centering
% \caption{Hyperparameters of PPO used for the experiments. The steps before PPO update are summed over all environments and $lin(x)$ is a linear schedule starting at $x$ and ending at $0$}
% \label{tab:hyperparameters}
% \begin{tabular}{c|c}
%     Hyperparameter & Value\\
% \hline
%     Total simulation steps & $5\times10^{5}$\\
%     Steps before update & $1024$\\
%     Minibatch size & $256$\\
%     Update epochs & $4$\\
%     Discount factor $\gamma$ & $0.995$\\
%     $\lambda_\text{GAE}$ & $0.95$\\
%     Clip range & $lin(0.1)$\\
%     Clip range value function & $0.2$\\
%     Value function coefficient & $0.5$\\
%     Entropy coefficient & $0.0$\\
%     Maximum gradient norm & $0.5$\\
%     Learning rate & $lin(2.5\times10^{-4})$
% \end{tabular}
% \end{table}

%The training is carried out for 500k steps using the main parameters shown in the table XX, please refer to the work [lapgym] and the code repository for further details. 
The training is carried out for $n_{iter} = 500000$ steps using the parameters reported in Table~\ref{tab:hyperparameters}, replicating those used in \cite{scheikl2023lapgym}. The public code repository also provides further details on parameters and implementation choices.
%using the same parameters as described in, please refer to the latter reference and the code repository for further details. The summary of used PPO parameters is presented in Table~\ref{tab:hyperparameters}.
Performances are measured by considering the overall training time and analyzing the rewards obtained.
All experiments are performed on a laptop with an Intel i7-9750HF CPU @ 2.60GHz, 16 GB RAM, and an NVIDIA  RTX 2060 Mobile GPU with 6GB VRAM. At the time of writing, this configuration is low-range and easily accessible to most researchers.

\section{RESULTS AND DISCUSSIONS}
All the data reported refers to the statistics calculated on 5 different executions, considering different initialization seeds.
In Table \ref{tab:fps}~(left), we report the performance of FF-SRL and the baseline considered (LapGym \cite{scheikl2023lapgym}), in terms of average and standard deviation of frames per second (fps) obtained as a function of the number of environments executed in parallel.
%In Table \ref{tab:fps} we report the performance, in the form of average and standard deviation of the fps obtained, of FF-SRL and of the baseline considered (Lapgym) as the number of environments executed in parallel varies. 
The first two columns show the performance considering only the simulation, while the others also consider the RL training. 
On the system configuration considered, FF-SRL is able to run up to 80 environments in parallel, providing a constant increase in the throughput, starting from the $2964.6\pm40.6$  obtained with a single environment, up to $23728.7\pm314$ with 64 environments, and then saturating the resources available at 80 environments with a result of $23932.7\pm317.4$, slightly above the configuration with 64 environments. We also ran FF-SRL on an NVIDIA A100 GPU and managed to run 1125 environments in parallel. This demonstrates how FF-SRL environment can scale even on high-performance computing systems.

For reference, the LapGym's simulation alone with a single environment is capable of achieving  $77.7\pm0.7$ fps, more than 38x slower than FF-SRL. A similar performance difference is also observed if we consider 4 and 8 environments running in parallel, where FF-SRL provides over 40x and 60x speedup, respectively. This is an expected result, since LapGym is based on the SOFA simulation framework, which relies on a bio-mechanical finite element model solved entirely on the CPU. 
%Considering instead only the simulation of a single environment in UnityFlexML, a simulator that takes advantage of the optimized PBD implementation provided by NVIDIA Flex, it manages to obtain $56.34\pm4.18$. This result is better than Lapgym, but far from the over 180 fps obtained by FF-SRL, mainly due to the overhead introduced by the Unity environment.

When evaluating pure simulation performance, it is interesting to analyze the fps of a single environment as the complexity of the biomechanical simulation varies, as reported in Table \ref{tab:fps}~(right).
In the second row, we can see how an increase of approximately 20\% in the number of vertices of the tissue model leads to a reduction in fps of about 5\% for both simulators. The situation changes starting from the third row, in which a model with approximately 2.5 times the vertices of the initial one leads to a reduction in fps of around 8\% for FF-SRL, while for Lapgym the reduction is around 40\%. The last two rows of the tables confirm the capability of FF-SRL to manage models with an even higher number of vertices, i.e. 9729 and 52359, respectively, unlike Lapgym, which sees performance halved and even collapses below 10 fps in the case of the most complex model considered. This result demonstrates how the optimized XPBD implementation on which FF-SRL is based can optimally handle the complex simulations of deformable structures required in RASS applications

% A rendering of the example setup with 64 parallel environments is presented in Fig.~\ref{fig:rendering}. To the best of our knowledge, there is no work in the field of simulation that supports the training of autonomous surgical systems capable of guaranteeing the performance achieved by FF-SRL.

% \begin{figure}
%     \centering
%     \includegraphics[width=\columnwidth]{images/multipleEnvs.png}
%     \caption{A FF-SRL simulation with 64 simplified tissue retraction environments running independently in parallel.}
%     \label{fig:rendering}
% \end{figure}

Considering the performance in terms of RL training, we report in the third and fourth columns of Table \ref{tab:fps}~(left), the statistics on the fps obtained from FF-SRL and Lapgym, respectively. 
Even in this context, we can observe how FF-SRL obtains a higher fps throughput than LapGym, providing significant performance increases. If we consider a single environment, FF-SRL yields a performance increase of over 6x with respect to Lapgym. %comparable to that obtained compared to UnityFlexML ($51.54 \pm3.88$, not shown in the Table).
If we consider 4 environments, we have an increase of more than 8.5x, which becomes over 13.5x with 8 environments, the maximum number of environments supported by LapGym on the hardware configuration considered. However, FF-SRL can also run 16 and 32 environments in parallel, providing a further increase in performance compared to the configuration of 8 Lapgym environments, of 20 and 26 times respectively. We should note that we were not able to run more than 32 FF-SRL environments during RL training on the considered system. This reduction compared to the simulation alone, capable of reaching 80 environments, is linked to the limits of the graphics memory.
%The GPU memory occupied by the simple simulation environment considered varies between 1.13~GB for the single environment up to 1.56~GB for the case with 32 environments.
The limit of 80 simulation environments in parallel is rather connected to the available system memory, since instantiating the simulation requires creating large arrays as described in Section \ref{sec:rl_games}. It will certainly be an element that we will improve, optimizing the data types used and memory management by limiting the initial copies of data between Warp and Pytorch.

As seen in Table \ref{tab:fps}~(left), the FPS reduction from simulation only to full RL pipeline is significant in FF-SRL as the simulation steps take only small fraction of the time needed for full RL processing. For Lapgym the simulation steps time is a major contribution to the RL processing time and therefore the reduction is less pronounced.

\begin{figure}[tb]
    \centering
    \includegraphics[width=0.95\columnwidth]{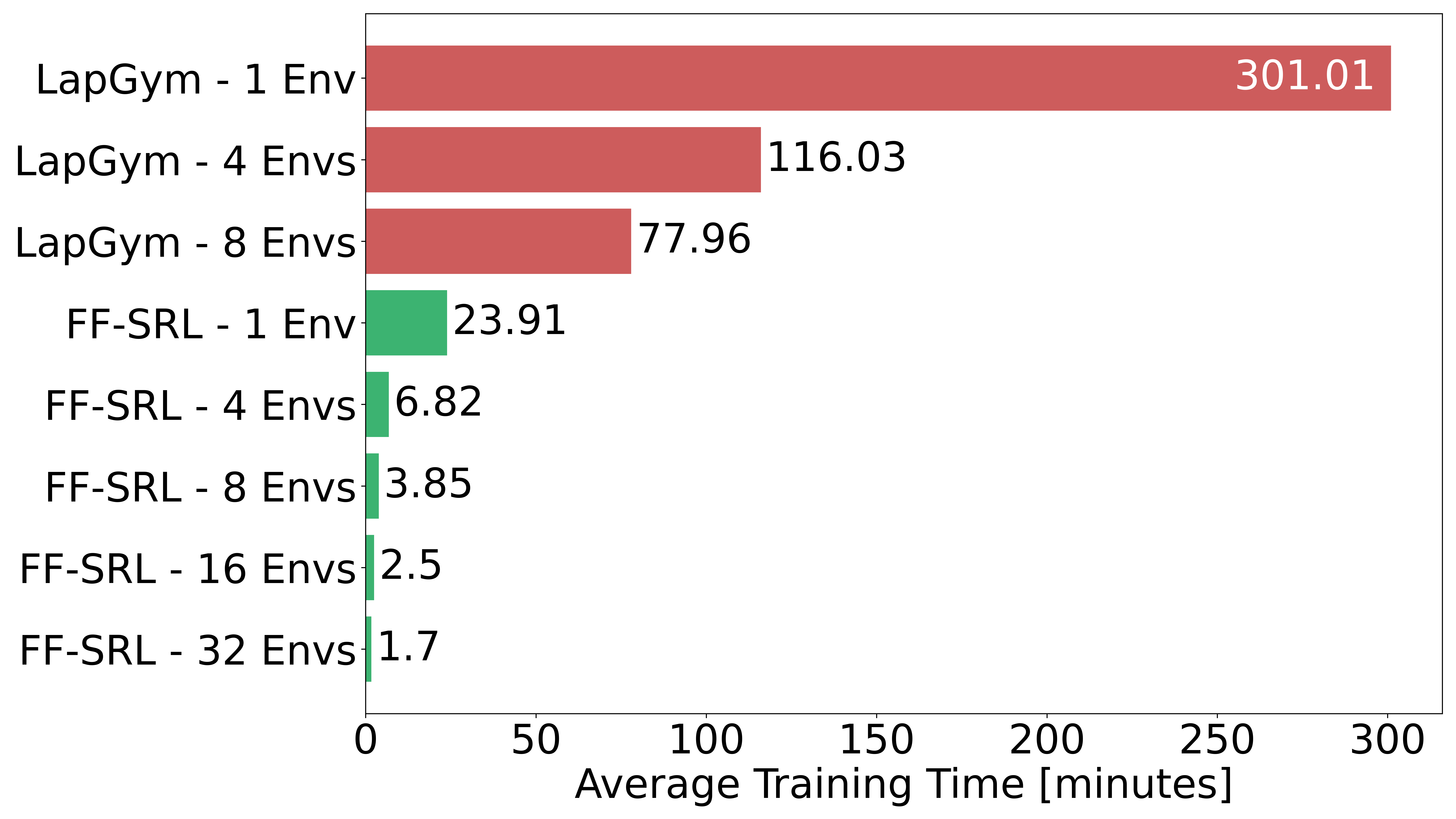}
    \caption{Average training time for $n_{iter} = 500000$ steps, considering different number of environment running in parallel. }
    \label{fig:timebar}
\end{figure}

Considering the overall RL training times shown in Fig.~\ref{fig:timebar}, the proposed simulator with a single environment completes training in less than 24 minutes compared to over 5 hours for the baseline. If we consider the maximum possible performances, we can complete the training in less than 2 minutes (32 environments configurations), guaranteeing a speedup of over 45 times compared to the 8 LapGym environment condition. Even comparing the 8 FF-SRL environments with the equivalent LapGym number, the speedup is still more than 20 times. It is important to observe how FF-SRL by going from 1 to 4 environments allows to reduce the training time by over 3.5 times, while LapGym achieves a reduction of approximately 2.6 times. Comparing the transition from 4 to 8 environments, FF-SRL obtains a time reduction of approximately 44\%, compared to approximately 33\% obtained by LapGym. These results confirm the scalability of FF-SRL with respect to the number of parallel environments.

\begin{figure}[tb]
    \centering
    \includegraphics[width=0.95\columnwidth]{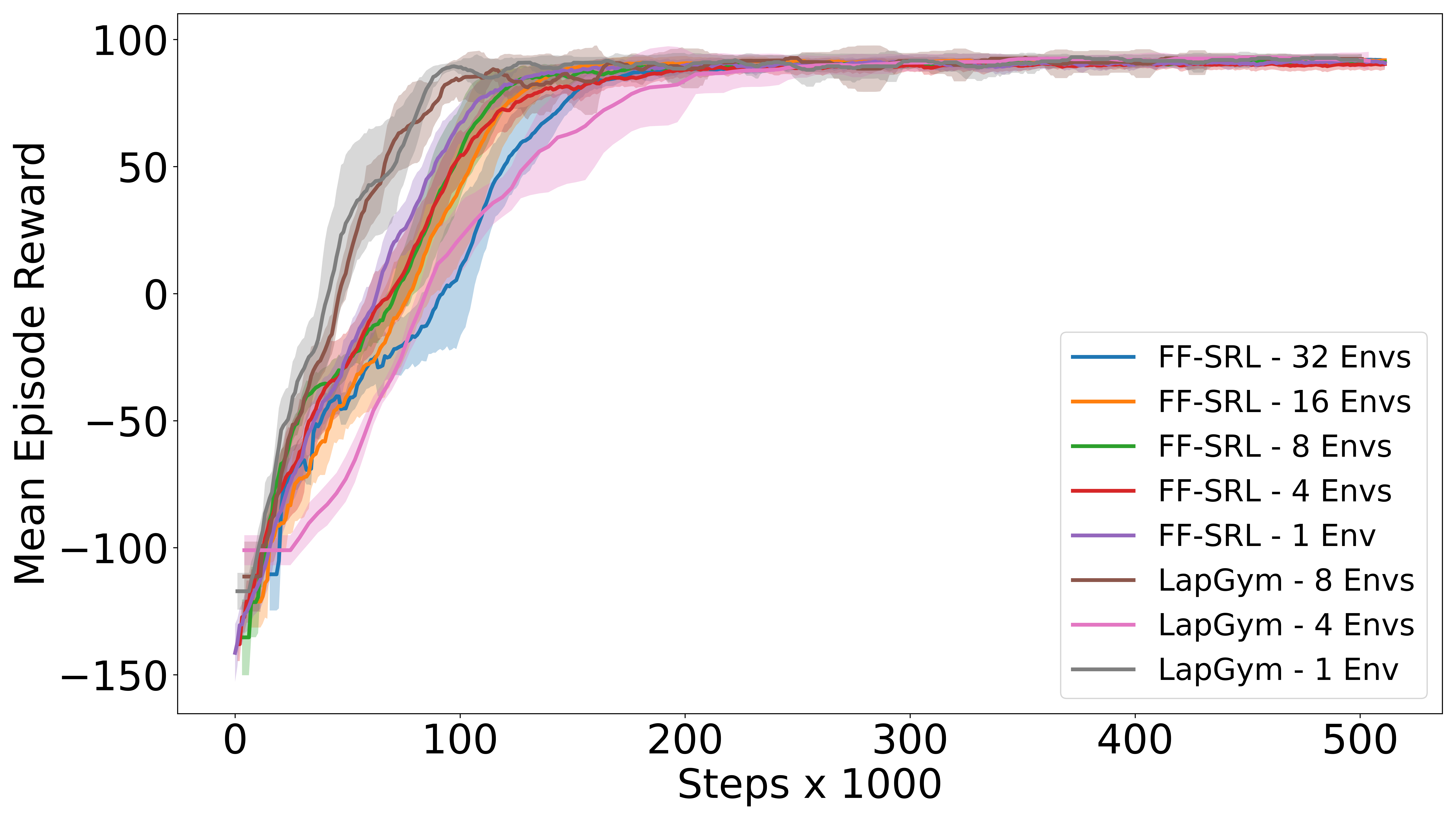}
    \caption{Average episode reward with respect to steps. }
    \label{fig:reward}
\end{figure}

Fig.~\ref{fig:reward} shows the curves of the average reward obtained per episode compared to the total number of steps. All the configurations considered are able to train an RL agent capable of solving the task considered (corresponding to a reward greater than 80). 
The convergence of RL training in FF-SRL and Lapgym is closely overlapping, guaranteeing a correct solution of the task already after 250,000 steps. This fact allows us to observe that the times reported in Fig.~\ref{fig:timebar}, which consider all 500,000 training steps, are overestimated and probably about half the time is sufficient to solve the considered task.
%It is possible to notice how the convergence of the LapGym environment is faster even if characterized by accentuated oscillations in the first 100k steps. This faster convergence may be due to some initialization aspects that could not be fully replicated (since LapGym uses StableBaselines3 \cite{raffin2021stable} as RL framework) and which lead to a more negative initial reward in the FF-SRL case.

In this work, we did not optimize the training hyperparameters to ensure a fair comparison with the baseline. 
Optimization of these parameters could favor a further increase in performance and allow the management of even more complex tasks, such as complete tissue retraction or tasks that require the use of two instruments.
These tasks have already been demonstrated to be able to be transferred effectively between simulator and reality~\cite{tagliabue2020unityflexml, Sheikl_SimToReal_2022,pore2021LFD}. Tests with the real robot will be carried out in the future to demonstrate the capabilities of FF-SRL to also interface with real robotic systems, exploiting ROS middleware.
%This, however, leads to a slower convergence of the reward as the number of environments increases, as already reported in \cite{makoviychuk2021isaac}. A careful setting of the hyperparameters (for example batch size and horizon length) to optimize the training process would certainly have allowed this aspect to be improved. 
%The extension of the study in this direction will be the focus of the upcoming research activities, together with 
In the future, we will also focus on the optimization of some technical aspects, such as improving the implementation of inverse kinematics to support all degrees of freedom of the dVRK and extending the evaluation to visual RL methods. 

\section{CONCLUSION}

% We have developed a framework that empowers the development of high-performance surgical training platforms. The simulator utilizes parallel calculation capabilities of GPUs and provides a skeleton for simulation of basic surgical tasks.

% Fig.~\ref{fig:averageFPS} demonstrates the robustness of the proposed framework. We have the capability to concurrently simulate dozens of environments in parallel and with increasing the number of environments we are increasing the effective fps performance up to the peak device performance around 200 environments.

% We firmly believe that this solution holds significant promise in the development of efficient RL policies for robot control within the context of surgical procedures. 

% Our upcoming research will concentrate on integrating the framework into RL policy learning algorithms and including differentiability to our simulations, with the goal of further improving policy training times.

% initially high because of high cpu overhead on creating scene, arrays, etc.

In this paper, we presented FF-SRL, a GPU-based simulation environment for robotic surgery that leverages an advanced XPBD simulation of deformable tissue. We showed that FF-SRL can significantly speed up the RL training process for surgical tasks, achieving higher frame-rates and faster training time than other available simulators. We also demonstrated the scalability and efficiency of our simulation environment, which can run on a single low-end GPU device. 
%Our work can make the use of RL techniques to support autonomous RASS systems more accessible, allowing to speed up research or consider more complex surgical tasks that include interaction with deformable tissues.
Our work opens up new possibilities for developing autonomous surgical systems using RL techniques, as well as for studying the interaction between robots and soft tissue. We hope that our code and simulator will be useful for the research community and foster further advances in this field.

\section*{ACKNOWLEDGMENT}
The authors want to thank Paul Maria Scheikl from Friedrich-Alexander-Universität Erlangen-Nürnberg and Miles Macklin from Nvidia for valuable support.
%This research is supported by the European Union’s Horizon 2020 research and innovation programme under grant agreement Sano no 857533, and by the International Research Agendas programme of the Foundation for Polish Science, co-financed by the European Union under the European Regional Development Fund.

\bibliographystyle{IEEEtran}
\bibliography{references}

\end{document}